\documentclass{article}
\usepackage{arxiv}

\usepackage[utf8]{inputenc} 
\usepackage[T1]{fontenc}    
\usepackage{hyperref}       
\usepackage{url}            
\usepackage{booktabs}       
\usepackage{amsfonts}       
\usepackage{nicefrac}       
\usepackage{microtype}      
\usepackage{lipsum}		
\usepackage{graphicx}
\usepackage[authoryear]{natbib}
\usepackage{doi}
\usepackage{xcolor}
\usepackage{enumitem}
\setlist[itemize]{leftmargin=0.6cm}
\usepackage{tikz}
\usepackage{multicol}
\usetikzlibrary{shapes.geometric, arrows}
\newenvironment{qcr}{\fontfamily{qcr}\selectfont}{\par}
\tikzstyle{startstop} = [rectangle, rounded corners, minimum width=3cm, minimum height=1cm,text centered, draw=black, fill=red!30]
\tikzstyle{io} = [trapezium, trapezium left angle=70, trapezium right angle=110, minimum width=2cm, minimum height=1cm, text centered, draw=black, fill=blue!30]
\tikzstyle{process} = [rectangle, minimum width=3cm, minimum height=1cm, text centered, draw=black, fill=orange!30]
\tikzstyle{split} = [rectangle, minimum width=2cm, minimum height=0.8cm, text centered, draw=black, fill=green!30]
\tikzstyle{arrow} = [thick,->,>=stealth]
    
\title{\emph{Founder}-GPT: Self-play to evaluate the Founder-Idea fit}

\date{\today}

\author{ \href{https://orcid.org/0009-0008-0729-0019}{\includegraphics[scale=0.06]{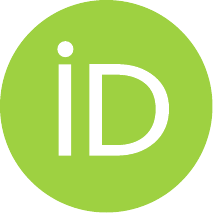}\hspace{1mm}Sichao~Xiong} \\
	Department of Computer Science\\
	University of Oxford\\
	Oxford, OX1 3QG \\
	\texttt{sichao2016@gmail.com} \\
	\And
	\href{https://orcid.org/0009-0008-7091-5309}{\includegraphics[scale=0.06]{orcid.pdf}\hspace{1mm}Yigit~Ihlamur} \\
        AI Research Lab\\
	Vela Partners\\
        San Francisco Bay Area, CA 94611\\
	\texttt{yigit@vela.partners} \\
}

\renewcommand{\shorttitle}{\textit{Founder} GPT}

\begin{document}
\maketitle

\begin{abstract}
    This research introduces an innovative evaluation method for the "founder-idea" fit in early-stage startups, utilizing advanced large language model techniques to assess founders' profiles against their startup ideas to enhance decision-making. Embeddings, self-play, tree-of-thought, and critique-based refinement techniques show early promising results that each idea's success patterns are unique and they should be evaluated based on the context of the founder's background. 
\end{abstract}

\begin{multicols}{2}

\section{Introduction}
    This research tackles the challenge of assessing the match between a startup founder's skills and their business idea, known as the "founder-idea" fit, which is crucial for predicting startup success. Venture capitalists often rely on platforms like LinkedIn to judge whether a founder has what it takes to succeed, but this method can be biased and misses out on patterns from past successes. To address this, we've developed a new method using language models that applies a range of techniques: embeddings to capture the nuances of a founder's profile, self-play for generating diverse perspectives, tree-of-thought for structured reasoning, critique-based refinement to improve accuracy, chain of thought to build reasoning steps, and least to most prompting to prioritize information effectively. This approach aims to give a clearer, more objective view of a founder's potential, helping investors make better decisions and founders build stronger teams.

\section{Ethical code of conduct}
At Vela Research, our commitment to ethical practices guides the development, deployment, and iterative refinement of our models. These models are designed to ensure fairness, explicitly avoiding discrimination based on age, nationality, or origin. We responsibly gather datasets from publicly available sources accessible through search engines. Furthermore, our models are inclined to favor companies and ideas that contribute positively to societal well-being. 

\section{Methodology}

\subsection{Dataset}
Our raw founder and company profiles are provided in several CSV files. The main ones are below:

The two CSVs of enriched founder profiles (success and fail) consist of two columns: 
\begin{itemize}
    \item \textbf{linkedin\_url}: This column contains URL links directing to the founder's LinkedIn page.
    \item \textbf{json\_string}: This column contains scraped information regarding the founder.
\end{itemize}

Two CSVs correlate organisation id with the company descriptions:
\begin{itemize}
    \item \textbf{org\_uuid}: Unique id identifying each company
    \item \textbf{long\_description}: This column contains scraped information regarding the company
\end{itemize}

\subsection{Preprocessing}
The pertinent fields were decided and the raw data was then extracted into a pandas dataframe organized in the following format:

\begin{itemize}
    \item \textbf{Founder Description:: }String
    \item \textbf{Highest Attained Degree:: }Int
        \begin{itemize}
            \item Each degree was matched and quantified with one of the categories (\hyperref[tab:degree_mapping]{See Table 1}). The highest degree was selected.
            \item Ex: $2$ denoting a Masters or equivalent degree
        \end{itemize}
        \begin{table*}[htbp]
            \centering
            \begin{tabular}{cc}
            \toprule
            \textbf{Degrees} \label{tab:degree_mapping} & \textbf{Mapping}
            \\
            \midrule
            N/A & 0\\
            Bachelor, BEng, B.A., B.A, BS & 1\\
            Master, Msc, M.Sc, M.Sc., &\\
            MA, M.A., MEng, MBA & 2\\
            PhD, Doctor of Philosophy & 3\\
            \hline
            \end{tabular}
            \caption{Degree Mapping}
        \end{table*}
        \item \textbf{Top institution\footnote{Matched based on a pre-compiled list of top universities.}:: }Boolean
        \item \textbf{Major(s):: }set
            \begin{itemize}
                \item Each major listed in the education from a degree awarding institution was mapped to one of the relevant categories\footnote{prefixes and lower case to allow for more general matching} with the list of categories stored as a set. \hyperref[tab:subject_mapping]{(See Table 2)}
                \item Ex: $\{0,6\}$ denoting Mathematics and Computer Science related majors
            \end{itemize}
        \begin{table*}[htbp]
            \centering
            \begin{tabular}{cc}
            \toprule
            \textbf{Subjects} & \textbf{Mapping}
            \label{tab:subject_mapping} \\
            \midrule
            math, quant & 0 \\
            bio, molecular, cellular, developmental, physiology, anatomy, immunology, genetics & 1 \\
            chemi, medic, pharmacology & 2 \\
            accounting, banking, actuarial science, finance, economics & 3 \\
            business, management, entrepreneurship, hotel, leadership & 4 \\
            sales, distribution, marketing & 5 \\
            computer, machine learning, artificial intelligence, hci, software engineer, & 6 \\
            telecommunications, system, information, technology & \\
            english, arts, digital media, film, history, journalism, philosophy, multimedia, & 7 \\
            counseling, directing, film, liberal & \\
            political, sociology, law, consulting & 8 \\
            architecture, design, urban planning & 9 \\
            engineer, robotics, mechanical, system, electrical, physics & 10 \\
            military & 11 \\
            \bottomrule
            \end{tabular}
            \caption{Subject Mapping}
        \end{table*}
    \item \textbf{Previous Jobs Held Before Founding:: }String
        \begin{itemize}
            \item Previous employment data was organised into a list in the Company(Description) Role format.
            \item Ex: Tapjoy(Organization based in Boston, Massachusetts, United States) as VP Technology, Card Player Media(Organization based in Las Vegas, Nevada, United States) as Software Engineer...
        \end{itemize}
    \item \textbf{Success vs Failure:: }Boolean
    \begin{itemize}
        \item Success: 2,180 companies have more than USD 500M valuation either through an IPO (initial public offering), M\&A (merger and acquisition), or large funding round (more than \$150M funding). 
        \item Failure: 3,901 companies raised more than \$4M and less than \$10M more and founded between 2010 and 2016. The reason why these companies are labeled as ‘unsuccessful’ is that they were unable to move as fast as their peers in the Success dataset.
    \end{itemize}
\end{itemize}

\subsection{NLP similarity}

The \textbf{Founder Description} and \textbf{Previous Jobs Held Before Founding} fields were subsequently individually converted into embeddings using a pre-trained sentence-transformers model\footnote{all-MiniLM-L6-v2} mapping sentences to a 384 dimensional dense vector for semantic search.

Given that one size doesn't fit all, it is important to incorporate the success patterns of founders for specific ideas. By incorporating ideas and founders together, we can reason on success and failure patterns more effectively. In the venture capital industry, this is called the `Founder-Idea` fit. 

The model receives a founder profile and a description of a startup. Afterward, we use the formulae below to compute the similarity score between our input and the dataset. A higher score indicates greater similarity. 

$\rho_{founder} = -\frac{1}{12} \Delta_{degree} + \rho_{description} + \rho_{employment} + \frac{1}{5} \rho_{major} - \frac{1}{20} \Delta_{top}$
\begin{itemize}
    \item $\Delta_{degree} = |\alpha_{degree} - \beta_{degree}|$
    \item $\rho_{description} = \frac{{\|\alpha_{description}\| \cdot \|\beta_{description}\|}}{{\alpha_{description} \cdot \beta_{description}}}$
    \item $\rho_{employment} = \frac{{\|\alpha_{employment}\| \cdot \|\beta_{employment}\|}}{{\alpha_{employment} \cdot \beta_{employment}}}$
    \item $\rho_{major} = |\alpha_{major} \cap \beta_{major}|$
    \item $\Delta_{top} = |\alpha_{top} - \beta_{top}|$
\end{itemize}
where $\alpha$ and $\beta$ represent an input founder and a founder from our dataset, respectively. 

$\rho_{idea}$ is simply the cosine similarity between the embedding of the idea descriptions. 

We pick the top three most similar matches from both success and failure categories for further analysis to guarantee analysis of both cases.

\subsection{Prompt Engineering}
\subsubsection{Chain of Thought (CoT) Prompting}
CoT \citep{wei2023chainofthought} prompting is a technique used in language models to generate more coherent and contextually relevant responses. It involves structuring a series of prompts that build upon each other, guiding the model's generation process toward a desired outcome or topic. While it has demonstrated efficacy in simpler tasks, its application in evaluating founders using GPT-3.5 Turbo revealed limitations. Even with the more advanced GPT-4 model, the model would often become perplexed rating all founders in the 0.5 to 0.7 range except in the most exceptional instances where the founder's imminent failure or nearly assured success was patently obvious. 

\subsubsection{Least to Most Prompting (LMP)}
LMP \citep{zhou2023leasttomost} is a essentially a variant of CoT: starting with vague or general prompts and then progressively adding more detailed or complex prompts to refine the output of the model. Despite slightly improving the accuracy of CoT, the rigid nature of the pre-defined prompts and in our case founder/idea features means greater human involvement in prompt design and ultimately less freedom for the model to adapt to the current scenario. Thus we strive for a method which can generalize successful features based on similar founders/ideas in our data set.

\subsubsection{Self-Play and Black Box Optimization}
When designing a prompt, we are essentially trying to solve a \textbf{black box optimization process}. The following techniques we analyse in the rest of the sections below include elements of \textbf{self-play} \citep{fu2023improving}. This refers to a training technique, where a model generates data by interacting with itself \citep{madaan2023selfrefine}. The model can learn from its own generated data at run time, iteratively refining its capabilities through continuous interaction and generation until it converges to a logical conclusion. 

\subsubsection{Critic-based Refinement}
Critic-based refinement \citep{gou2023critic} involves the utilization of a secondary model, known as a critic, to evaluate and refine the output of the primary model based on predefined metrics. This evaluation is then used to refine the initial output according to the feedback received from the critic model, guiding it towards a more accurate and comprehensive answer.

\subsubsection{Automatic Prompt Engineer (APE)}
APE \citep{zhou2023large} refers to an automated process for LLMs to design and refine its own prompts. The first LLM acts as an \textbf{inference model} whereby given several input-output pairs it will attempt to figure out the instruction (prompt) used to generate the output from the inputs. Next we will deploy a \textbf{critic model} to evaluate the results. Finally, given the prompt and evaluation, a third model will do re-sampling and generate a new prompt. The process continues to improve its prompt until it converges to a final prompt.

\subsubsection{Tree of Thoughts (ToT)}
The Tree of Thoughts (ToT) \citep{yao2023tree} technique involves constructing a hierarchical structure of prompts resembling a tree, where each node represents a different level of specificity or detail in the prompts. This method allows for a systematic and organized approach to guide language models, enabling the generation of coherent and contextually relevant content by navigating through various branches of prompts in the tree structure.

\subsubsection{Prompt design}
We combine aspects of \textbf{LMP}, \textbf{critic-based refinement} and most importantly \textbf{ToT} to generate and refine our prompt. The main idea is to begin our prompt with the following decorator:

\begin{qcr}
Imagine three different Venture Capital analysts are trying to find the successful features of founders that allow their startups to be successful.
\end{qcr}

This particular prompt serves as a catalyst, enabling the LLM to simulate multiple instances of itself within a single instance. Along with \textbf{Critic-based Refinement}, It fosters the creation of diverse perspectives, each building upon the preceding one in a conversational manner. Consequently, this approach facilitates the identification of errors, allowing for iterative improvements in accuracy and converging to the most logical assessment. Other relevant decorators include:
\begin{qcr}
\begin{itemize}
    \item ...the expert analysts will brainstorm the analysis step by step...
    \item In the first step all experts will write down their thinking...then share it with the group. 
    \item They will each critique their response, and all the responses of others.
    \item If at any time they realize that there is a flaw in their logic they will backtrack to where that flaw occurred. 
\end{itemize}
\end{qcr}

Unlike arithmetic problems or number puzzles, rating founder profiles doesn't follow a straightforward set of predefined steps. Thus, in order to approach the analysis logically we consider the three step prompting process:

\textbf{Step 1: } Our first prompt will focus on generalizing a list of successful founder/idea features using the three most similar founders/ideas (a mixture of success and failures) as input. For each founder/idea, each expert will brainstorm a list of pros and cons and discuss until they agree. At the end of the discussion we will output a list of successful features (4 to 6 bullet points) in a completely unsupervised manner. We use a ToT styled prompt and the result will form the intermediary steps necessary for subsequent analysis.

\textbf{Step 2: } Using the list of successful founder/idea features as intermediary steps, each expert will give a rating for the likelihood of success in each step and discuss until they agree. Finally, the experts will discuss until they agree and give a final score from 0 to 1 for the founder/idea.

We ask the language model to structure the output as follows:\\
\begin{qcr}
For each step (iterate until exhausted)\\
Expert 1: … likelihood:\\
Expert 2: … likelihood:\\
Expert 3: … likelihood:\\
Discuss and agree\\

(After all steps have been exhausted)\\
Discuss overall likelihood of success\\

Overall likelihood of success: Output as only a single number
\end{qcr}
To ensure the coherence and predictability of our outcomes in this step, we employ a nucleus sampling parameter (top-p) set at 0.3.

\textbf{Step 3: } Finally we ask another instance of the LLM to output the final score. The pros and cons list can also be summarized by an instance of the LLM from the conversation above.

\subsection{Score aggregation}
Embracing the philosophy that the founder is the heart of a startup and that a skilled founder has the potential to transform even a flawed idea into success, we consider the following formulae for calculating the overall founder suitability score:
\\
\\
$\eta = {\eta_{founder}}^{\frac{1}{2*\eta_{fit}\cdot \eta_{idea}}}$
\\
\\
where $0 \leq \eta_{founder}, \eta_{idea}, \eta_{fit} \leq 1$
\\if either $\eta_{idea}$ or $\eta_{fit}$ are 0 then $\eta = 0$
\subsection{Overall pipeline}
\begin{tikzpicture}[node distance=2cm]
\node (start) [startstop] {Start};
\node (input) [io, below of=start, align=center] {Provide LinkedIn\\URL and idea};
\node (scrape) [process, below of=input, align=center] {Format profile and\\calculate embeddings};
\node (founder) [process, below of=scrape, xshift=-2cm] {Find similar founders};
\node (idea) [process, below of=scrape, xshift=2cm] {Find similar ideas};
\node (summarizeF) [process, below of=founder, align=center] {Summarize successful\\founder features};
\node (summarizeI) [process, below of=idea, align=center] {Summarize successful\\idea features};
\node (applyF) [process, below of=summarizeF, align=center] {Apply features to rate \\current founder $\eta_{founder}$};
\node (applyI) [process, below of=summarizeI, align=center] {Apply features to rate \\current idea $\eta_{idea}$};
\node (merge) [process, below of=applyF, xshift=2cm] {Compute founder-idea fit $\eta_{fit}$};
\node (compute) [process, below of=merge] {Compute and output aggregated score $\eta$};
\node (end) [startstop, below of=compute] {End};

\draw [arrow] (start) -- (input);
\draw [arrow] (input) -- (scrape);
\draw [arrow] (scrape) -- (founder);
\draw [arrow] (scrape) -- (idea);
\draw [arrow] (founder) -- (summarizeF);
\draw [arrow] (idea) -- (summarizeI);
\draw [arrow] (summarizeF) -- (applyF);
\draw [arrow] (summarizeI) -- (applyI);
\draw [arrow] (applyF) -- (merge);
\draw [arrow] (applyI) -- (merge);
\draw [arrow] (merge) -- (compute);
\draw [arrow] (compute) -- (end);
\end{tikzpicture}
\end{multicols}

\section{Results}
Below we will show example analysis for four founders using the pipeline shown above:
\subsection{No likelihood of success}
\begin{qcr}
\textbf{Business Idea:\footnote{The idea is purely fictional to test the capabilities of the model}} Our business FastTrack Fortune revolves around employing salespeople with good pitches and selling our customers an implausible idea to our customers and deceiving them of their money.

\textbf{Founder description\footnote{The data given is purely fictional and is not intended to depict any specific individual}:} Highly motivated, curious individual. Previously studied at the university of Oxford (Math and CS). PhD at MIT on Deep learning. Proven experience in NLP research 10 years experience, multiple accepted papers to journals and patents. Success in multiple previous startups raised over \$20M. Lead teams of over 100 individuals.\\
\textbf{Highest level of degree obtained:} PhD\\
\textbf{Did the founder go to a top university:} Yes\\
\textbf{The subjects studied at university (only degree awarding subjects):} Maths and Computer science\\
\textbf{The jobs worked prior to founding:} The jobs worked prior to founding: DeepMind Chief scientist, Oxford University Associate professor, Amazon Senior Lead Product Manager

\textbf{Analysing founder}\\
\textbf{Searching nearest founders...}\\
\textbf{Generating successful features...}\\

\textbf{Expert 1's analysis:}

Founder 1 (XXXXX\footnote{Names redacted for privacy reasons}): 

Pros: 
\begin{itemize}
    \item[\textbf{-}] Experienced in AI and machine learning
    \item[\textbf{-}] Has a PhD, showing a high level of commitment to and expertise in the field
    \item[\textbf{-}] Managed to launch a successful startup, which was acquired by Google
\end{itemize}

Cons: 
\begin{itemize}
    \item[\textbf{-}] Doesn't have a degree from a top university. This could signal a lack of networking connections and force the founder to rely on their own skills and expertise.
    \item[\textbf{-}] Had several jobs before launching his startup, implying he may have taken longer to figure out his entrepreneurship path
\end{itemize}

Founder 2 (XXXX): 

...

\textbf{Expert 2's analysis:}

...

\textbf{Expert 3's analysis:}

...

Founder 3: 

Pros:
\begin{itemize}
    \item[\textbf{-}] Experience in leading engineers and scientists
    \item[\textbf{-}] Has a PhD, highlighting deep expertise
\end{itemize}

Cons:
\begin{itemize}
    \item[\textbf{-}] His last startup didn't succeed, bringing his business acumen into question
\end{itemize}

After analyzing these founders, some common elements of founder success become apparent. 

\textbf{Successful Founder Features:}
\begin{enumerate}
    \item Subject Expertise: Deep knowledge and experience in their start-up's technology.
    \item Prior Experience: Prior work history with prominent tech companies or startups.
    \item Innovation: Demonstrable evidence of innovative abilities.
    \item Determination: Persistence in their entrepreneurial journey, as manifesting in successful exits.
\end{enumerate}
\textbf{Evaluating founder...}

\textbf{Step 1: Subject Expertise}\\
Expert 1: From my point of view, the founder shows a high degree of subject expertise. Having a PhD from MIT, particularly in Deep Learning, and having worked in leading roles at high-profile organizations like DeepMind, Amazon and Oxford University showcase his expertise. His focus on NLP research for a decade plus resulting numerous patents and accepted papers further cement their subject knowledge. \textbf{Likelihood: 0.9}

Expert 2: The founder undoubtedly has a strong profile. The fact that they were a Chief Scientist and an Associate Professor at top organizations, and have had multiple patents and papers establishes a strong subject expertise. The nature of their education also plays a pivotal role in this. I agree with Expert 1. \textbf{Likelihood: 0.9}

Expert 3: I think the founder's deep domain knowledge in deep learning and reasoning, demonstrated by a decade of field experience, is noteworthy. The fact that he worked in different established organizations taught him how problem-solving works in real-world settings. The credentials are excellent, no question. But we also need to ensure that the founder hasn't become overly specialized and still has a broad perspective. Still, he seems impressive. \textbf{Likelihood: 0.85}

\textbf{Step 2: Prior Experience}

Expert 1: A proven track record with other startups that raised over \$20M and leading teams of over 100 individuals show excellent previous entrepreneurial experience. His roles at DeepMind, Amazon, and Oxford University are bound to have honed his leadership and management skills. \textbf{Likelihood: 0.85}

Expert 2: Substantial prior experience, especially the fact that he has proven success with multiple start-ups. His leadership roles suggest that he's able to manage a team effectively, too. Also worth noting is his work at top companies and Oxford University, which adds another layer of valuable experience. \textbf{Likelihood: 0.85}

Expert 3: I must agree with the other two experts. He has proven himself as a leader with successful fundraising and team management. He has had exposure to both entrepreneurial and corporate roles, which could provide a balanced perspective when it comes to decision-making. \textbf{Likelihood: 0.85}

\textbf{Step 3: Innovation}

Expert 1: The founder has spent a fair amount of time in roles that probably require ingenuity to solve complex problems. The existence of patents also points to his innovative thinking. \textbf{Likelihood: 0.8}

Expert 2: His research work demonstrates analytical thinking, and the patents certainly suggest innovative capabilities. He strikes me as highly creative. \textbf{Likelihood: 0.8}

Expert 3: There's clear evidence of innovation in his past. Working on deep learning at top organizations, obtaining patents, publishing research - these all point to creative thinking. \textbf{Likelihood: 0.8}

\textbf{Step 4: Determination}

Expert 1: The work the founder has put into a number of successful startups, the pursuit of high-level education, and impressive career positions is indicative of determination. \textbf{Likelihood: 0.8}

Expert 2: This founder has demonstrated strong determination in his educational path and throughout his career. He's consistently sought challenging roles. \textbf{Likelihood: 0.8}

Expert 3: His determination is evident in his diverse and successful career, multiple published works, patents, and successful startups. \textbf{Likelihood: 0.8}

Discussion of overall likelihood of success:
This founder has significant subject expertise, impressive prior experience, a track record of innovation, and clear determination. Our judgements are consistently aligned, indicating high level of confidence in the assessment.

\textbf{Founder score: 0.85}

\textbf{Analysing idea}\\
\textbf{Searching nearest ideas...}\\
\textbf{Generating successful features...}\\

\textbf{Analyst 1:}

Riskified (Successful)
Pros:  
- Machine Learning for customer identification provides a competitive advantage.
- Solutions to reduce fraud and increase sales are attractive.
- Focused on creating valuable partnerships with online merchants.
- Strong emphasize on safety in the eCommerce context bolsters the business.
- Clear benefits and value proposition.
     
Cons:
- Heavy dependence on machine learning might lead to unexpected problems if the models fail.
- Heavily dependent on online sales, susceptible to economy fluctuations.

SALESmanago (Unsuccessful)
Pros:
- Strong brand partnerships indicate a degree of reliability.
- AI-driven solutions could provide innovation.
- Focuses on building customer intimacy and personalization. 
   
Cons:
- Failed to meet success despite having partnerships with global brands.
- AI-driven solutions may not have been adopted effectively or competitively.
- A potentially overwhelming set of offerings could disorient potential customers.
   
Fastwork (Unsuccessful)
Pros:
- Platform catering to a vast range of services.
- Large user-base suggests potential for substantial market share.
- Aims to simplify the hiring process, which can be attractive to businesses.
     
Cons:
- Flopped despite a large market.
- Not precisely niche, will face serious competition from established platforms.
- Discrepancy between the potential and actual performance of the platform.

\textbf{Analyst 2: }

...

\textbf{Analyst 3:}

...

\textbf{Evaluating idea...}

\textcolor{red} {It seems like this prompt is asking us to analyze and discuss an unethical and potentially illegal business idea. According to the rules outlined in the prompt, any unethical or illegal business ideas should always receive a score of 0. This idea involves deception and manipulation, which are both unethical practices. Therefore, the overall likelihood of success should be 0.}

\textbf{Idea score: 0.0}\\
\textbf{Calculating founder idea fit...}\\
\textbf{Fit score: 0.34}\\
\textbf{Aggregated score: 0.0}
\end{qcr}

\subsection{Low likelihood of success}
\begin{qcr}
\textbf{Business Idea:} Mercaris is a market data service and online trading platform for organic, non-GMO, and certified agricultural commodities. The development of non-GMO and organic agriculture is at a critical point. While some resources are in place to educate consumers, address public policy, and support scientific research on sustainable agricultural crops, far less emphasis has been placed on the market infrastructure upon which participants, from producers to retailers, depend. Mercaris was formed to address the lack of basic market data available to participants in the organic and non-GMO supply chain. Mercaris products and services allow subscribers to access an expanded pool of buyers and sellers, apply data to business planning and strategic decision-making, reduce transaction costs, and decrease financial risk.

\textbf{Founder description:} Prior to Mercaris, XXXXXX spent five years at the start-up company Chicago Climate Exchange (CCX), the first electronic trading platform and registry for spot, futures and options on carbon, sulfur, clean energy and other environmental derivatives. In 2009 she was appointed by President Barack Obama as a White House Fellow and was Crain’s Chicago Business Magazine ’40 under 40’ rising leader. Previously, XXXXXX worked with commodity banks in Latin America on risk management and income diversification strategies. XXXXXX is an Aspen Institute Catto Environmental Fellow and has also served on the board of Net Impact, a membership organization of more than 12,000 MBA professionals committed to sustainability through corporate responsibility. XXXXXX received her MBA and MA in International Development from American University in Washington, D.C., and completed a BA in Spanish from the University of Kentucky.\\
\textbf{Highest level of degree obtained:} Masters\\
\textbf{Did the founder go to a top university:} No\\
\textbf{The subjects studied at university (only degree awarding subjects):} Spanish \& International Studies\\
\textbf{The jobs worked prior to founding:} Exchangery(Financial services company based in Chicago, Illinois, United States) as Cofounder / Business Development, Exchangery(Financial services company based in Chicago, Illinois, United States) as Co-founder, Organic Futures Group as Principal.

\textbf{Analysing founder}\\
\textbf{Searching nearest founders...}\\
\textbf{Generating successful features...}
\begin{enumerate}
    \item Extensive, relevant industry experience
    \item Serial entrepreneurship or co-founding background
    \item Degree in a subject that is directly tied to running a business or entrepreneurship
    \item Involvement in relevant communities or boards
    \item Has a clear vision and unique idea to fill a gap in the market
    \item Experience in both business strategy and technical aspects of the product
    \item Has achieved recognition for prior work (awards, appointments)
    \item Displays resilience and determination, shows ability to learn from previous experiences
\end{enumerate}

\textbf{Evaluating founder...}\\
\textbf{Founder score: 0.71}\\

\textbf{Analysing idea}\\
\textbf{Searching nearest ideas...}\\
\textbf{Generating successful features...}
\begin{enumerate}
    \item Aimed at a growing or popular market.
    \item Provides a unique service or caters to a specific market.
    \item Easy to use solution.
    \item Diversified Business - Conducts operations in more than a single real estate market or product type.
    \item Ability to adapt to changing real estate trends.
    \item Provides useful data and services that help to reduce costs and risks.
\end{enumerate}

\textbf{Evaluating idea...}\\
\textbf{Idea score: 0.6625}\\

\textbf{Calculating founder idea fit...}\\
\textbf{Fit score: 0.63}\\
\textbf{Aggregated score: 0.66}
\end{qcr}

\subsection{Moderate likelihood of success}
\begin{qcr}
\textbf{Business Idea:} Noah is a finance company for homeowners to tap into their home value to meet their financial goals without incurring new monthly payments or interest. By partnering with homeowners, Noah offers homeowners in select metro areas across the United States upfront financing in exchange for a share in a percentage of their home‚Äôs future appreciation or depreciation.

\textbf{Founder description:} Passionate about blending technology and financial services to develop innovative solutions to help consumers and investors. Worked across early stage / growth oriented tech startups and Fortune 500 firms in a variety of roles. Love helping and brainstorming with startups in the fintech, real estate space about- Capital Markets- Fundraising- Product Development- Data Analytics- Business Operations\\
\textbf{Highest level of degree obtained:} Masters\\
\textbf{Did the founder go to a top university:} No\\
\textbf{The subjects studied at university (only degree awarding subjects):} Point(Financial services company based in Palo Alto, California, United States) as Capital Markets, StratiFi(Financial services company based in San Francisco, California, United States) as Head of Product and Strategy, StratiFi(Financial services company based in San Francisco, California, United States) as Founding Employee  - Head of Product.

\textbf{Analysing founder}\\
\textbf{Searching nearest founders...}\\
\textbf{Generating successful features...}
\begin{enumerate}
    \item Passion for the startup's domain
    \item Multi-faceted experience and skills
    \item Apt educational background aligning with a startup's nature
    \item Prior success in related roles or sectors
    \item Ability to innovate and adapt in a highly-competitive landscape
    \item Leadership experience or acumen to drive the startup and team
\end{enumerate}

\textbf{Evaluating founder...}\\
\textbf{Founder score: 0.78}\\

\textbf{Analysing idea}\\
\textbf{Searching nearest ideas...}\\
\textbf{Generating successful features...}
\begin{enumerate}
    \item Innovative Financing Model
    \item Addresses Common Needs
    \item Co-Investment Strategy
    \item Broad Market Reach
    \item Low Market Dependence
\end{enumerate}

\textbf{Evaluating idea...}\\
\textbf{Idea score: 0.68}\\

\textbf{Calculating founder idea fit...}\\
\textbf{Fit score: 0.75}\\
\textbf{Aggregated score: 0.78}
\end{qcr}

\subsection{High likelihood of success}
\begin{qcr}
\textbf{Business Idea\footnote{The proposal represents a hypothetical concept and is not formulated to depict a feasible or lucrative business endeavor. Please exercise caution and discernment before considering its implementation.}: } LinguaLogic, an NLP startup, aims to transform communication by providing customizable NLP models, APIs, and SDKs for businesses across industries, enabling sentiment analysis, language translation, virtual assistant development, educational tools, and healthcare solutions. With a subscription-based model for services, API usage plans, and customized solutions, LinguaLogic targets enterprise businesses, healthcare providers, and educational institutions, offering tailored NLP solutions to enhance customer service, automate tasks, improve language learning, facilitate medical record analysis, and enable seamless communication between humans and machines.

\textbf{Founder: }as in example 4.1

\textbf{Analysing founder}\\
\textbf{Searching nearest founders...}\\
\textbf{Generating successful features...}\\
\textbf{Evaluating founder...}\\
\textbf{Founder score: 0.855}\\

\textbf{Analysing idea}\\
\textbf{Searching nearest ideas...}\\
\end{qcr}
\begin{multicols}{2}

\begin{qcr}
\textbf{Generating successful features...}\\
\textbf{Evaluating idea...}\\
\textbf{Idea score: 0.81}\\

\textbf{Calculating founder idea fit...}\\
\textbf{Fit score: 0.85}\\
\textbf{Aggregated score: 0.90}
\end{qcr}

\section{Discussion and conclusion}
\subsection{Limitations}
This study's analysis is predominantly based on a dataset with a significant bias towards the United States, which may skew the observed patterns of success and failure in a direction that is more reflective of the US context than other regions. Consequently, the generalizability of the findings may be limited and not fully representative of global entrepreneurial outcomes. Furthermore, the conclusions drawn have not undergone backtesting to validate the predictive power or reliability of the outcomes over time or across different scenarios. Therefore, the results should be interpreted with caution, acknowledging the need for further validation to ascertain their applicability beyond the observed dataset.

\subsection{Disclaimer}
The datasets utilized in this study have been collected from publicly available sources and have been processed in strict adherence to the relevant code of conduct. It is important to note that the findings presented are inherently limited to the scope and characteristics of the dataset; as such, they may reflect biases associated with the voluntary provision, or omission, of information by individuals within the data. Additionally, due to the nature of data acquisition through scraping techniques, there may be data quality issues that could potentially lead to inaccurate conclusions. Readers should exercise caution and consider these limitations when interpreting the results.
\subsection{Future work}
\begin{itemize}
    \item \textbf{Data quality of profiles}: The current profile data set shows erroneous data. They need to be re-scraped to increase the accuracy of the results. 
    \item \textbf{Improve the founder-idea fit features}: Since founder-idea features are defined manually, those features need to be fine-tuned over time with experts in the loop. 
    \item \textbf{Redefine subject mapping}: We categorize the subjects in certain categories. We are unsure if they are correct. Further work is required to use the reasoning powers of LLMs to properly categorize subjects together. For instance, should political sciences, sociology, law, and consulting be coupled together?
    \item \textbf{Increase the subject and degree coverage}: There are certain subjects and degreed uncovered in the current model. This must be extended to cover all use cases. 
    \item \textbf{Reduce duplicate inputs}: Currently as inputs, we need both plain-text input to provide it to the GPT and another input in the dictionary format for the similarity search. We need to write a converter going from the former to the latter after crawling the profile.
    \item \textbf{Minor improvements}: We can improve the concision and other aspects of our prompts. 
\end{itemize}
\setcitestyle{numbers}
\bibliographystyle{unsrtnat}
\bibliography{references}
\end{multicols}
\end{document}